\documentclass[letterpaper, 10 pt, conference]{ieeeconf}
\IEEEoverridecommandlockouts
\usepackage[T1]{fontenc}
\usepackage{amsmath,amssymb}
\usepackage{booktabs}
\usepackage{xspace}
\usepackage{listings}
\usepackage{xcolor}
\usepackage{cite}
\usepackage{url}
\usepackage{tikz}
\usetikzlibrary{positioning,arrows.meta,calc}
\usepackage{pgfplots}
\pgfplotsset{compat=1.18}
\usepgfplotslibrary{groupplots}
\definecolor{cE}{RGB}{31,102,190}
\definecolor{cDDM}{RGB}{22,150,110}
\definecolor{cDyn}{RGB}{214,90,140}
\definecolor{cF}{RGB}{60,60,60}

\newcommand{\dlt}{DELTA\xspace}
\newcommand{\s}{\mathbf{s}}
\newcommand{\shat}{\hat{\mathbf{s}}}
\newcommand{\uu}{\mathbf{u}}
\newcommand{\vv}{\mathbf{v}}
\newcommand{\e}{\mathbf{e}}
\newcommand{\dd}{\mathbf{d}}
\newcommand{\dyn}{DynSSM\textsuperscript{*}\xspace}
\newcommand{\dynp}{DynSSM\textsuperscript{*+}\xspace}
\newcommand{\ddmh}[1]{DDM$^{\text{*}}_{#1}$\xspace}
\newcommand{\ddmp}[1]{DDM$^{\text{*+}}_{#1}$\xspace}
\newcommand{\ddms}{DDM$^{\text{*}}$\xspace}
\newcommand{\ddmstep}{DDM-step\xspace}
\newcommand{\pc}{\,\%\xspace}

\begin{document}

\title{\LARGE \bf DELTA: Online Disturbance Learning for Cross-Track Vehicle Dynamics Prediction in Autonomous Racing}

\newif\ifanon \anonfalse
\ifanon
\author{Anonymous Authors}
\else
\author{Niklas Holle%
\thanks{N. Holle is with the University of Stuttgart, Stuttgart, Germany (e-mail: \texttt{st169603@stud.uni-stuttgart.de}).}}
\fi

\maketitle
\thispagestyle{empty}
\pagestyle{empty}

\begin{abstract}
In autonomous racing, vehicle dynamics models are usually developed and fitted on track data. Afterwards, they are used on the same track or on another one. The latter case is more complex, especially when the type of the track changes. We test three different models, fitted on a road course, on a banked oval: a single-track physics model, the Deep Dynamics Model and a re-implementation of DynSSM. On the road course, the physics model is more accurate on the 2\,s horizon than an extrapolation with constant velocities. This changes on the banked oval: here, all three are worse than the extrapolation. While driving, the car sees the error of each one-step prediction one sample later. We propose \dlt, an online layer which sorts these errors into a table over speed and lateral acceleration. Every prediction step adds the matching table value. It needs no offline training and about 30 lines of code. Furthermore, it is deployable with all three models we evaluate and lowers the 2\,s error of the physics model and DynSSM by 60 to 66\pc, that of DDM by 34 to 55\pc. With \dlt, the physics single-track model comes close to the best learned model, about 8\pc behind it at 2\,s when fitted from neutral initial values, and it predicts 16\,384 candidates over 2\,s in about 10\,ms on a GPU. The gain of \dlt is confirmed with data from another team and car, using the BETTY dataset. Using \dlt, all three models keep the car on track with improved predictions in a closed-loop simulation on unseen ovals.
\end{abstract}


\section{Introduction}
Operating a race car at the limits of handling with model predictive control (MPC) relies on accurate vehicle models whose predictions hold up over the whole horizon. In full-scale competitions such as the Indy Autonomous Challenge (IAC), this horizon can reach 2.6\,s~\cite{raji2023}, while learned models are evaluated over 0.6\,s~\cite{chrosniak2024,dynssm} to 1.7\,s~\cite{dkmgp}. Cheap models are needed for sampling-based controllers, which query them for up to thousands of candidate inputs per control cycle~\cite{williams2017}. Finally, race cars are deployed on many different tracks~\cite{betz2022}, so the model must work on unseen tracks.

Learned and hybrid models, which are trained on the target vehicle, match or outperform physics models~\cite{spielberg2019}, satisfying the first requirement. Deep Dynamics (DDM) estimates single-track coefficients, including those for a Pacejka tire model, inside physical bounds with a network~\cite{chrosniak2024}. DynSSM uses these bounded parameters and combines them with a temporal memory and an additional residual correction~\cite{dynssm}. Both models are trained on a road course (Putnam Park) and tested on a banked oval (Las Vegas Motor Speedway, LVMS) using the recordings attributed to~\cite{racecar} of an IAC car.

We noticed two things while reproducing these benchmarks. Simply holding the current velocities constant results in a position error after 0.6\,s that is only 2\,cm larger than that of a simple oracle which integrates the measured future velocities (DDM protocol; 5\,cm in the DynSSM protocol), so this horizon is rather short. At 2\,s, this looks different. The constant-velocity extrapolation is ahead of all three models (single-track physics model, DDM and our re-implementation of DynSSM, called \dyn) (Table~\ref{tab:main}). Fitted only on a road course, these models face differences in speed, tire loads and banking on the oval.

During the run, the car experiences these differences directly. One sample after each one-step prediction, it can be compared to the measurement. \dlt (Delayed-Error Learning for Trajectory Adaptation) works with this delayed error. The errors collected during driving fill a small table over speed and lateral acceleration. During a prediction, each step is corrected with its matching table value. \dlt can be combined with every model we evaluate, as it requires one-step predictions only. Our contributions are:

\begin{enumerate}
    \item a study of the Putnam-to-LVMS transfer at two horizons (0.6\,s and 2\,s) with additional reference predictors (constant-velocity extrapolation and a simple oracle);
    \item \dlt, an online layer of roughly 30 lines of code, which cuts the 2\,s error of the physics model and \dyn by 60 to 66\pc and DDM by 34 to 55\pc;
    \item a confirmation on a second IAC dataset and car and closed-loop tests with a sampling-based MPC in simulation.
\end{enumerate}

Minimal reference implementations of the physics host with \dlt (results and computation time) and of \dyn with and without \dlt are available at \url{https://github.com/niklas-holle/DELTA/tree/preprint-2026-11}.

\section{Related Work}

\emph{Learned and hybrid vehicle models.} Besides the hybrid models DDM and DynSSM, Gaussian-process based models with deep kernels were published for full-scale race cars, e.g., DKL-SKIP~\cite{dklskip} and DKMGP~\cite{dkmgp}. All four are trained offline on datasets; DDM and DynSSM re-estimate their coefficients from the latest history, but with fixed weights. \dlt can be added to such a model to correct its error while driving.

\emph{Online adaptation.} Kabzan et~al.~\cite{kabzan2019} use a sparse Gaussian process, updated during the run, to correct a nominal model. Other approaches are the improvement of Gaussian-process corrections in between runs of a simulator~\cite{su2023} or the selection of the best physics model from a bank of candidates, as in LLA-MPC~\cite{llampc}. \dlt combines two older ideas for multi-step prediction: additive disturbances from prediction errors estimated in offset-free MPC~\cite{pannocchia2003,maeder2009} and look-up tables of engine control adapted online~\cite{vogt2004}.

To our knowledge, three methods are closest to \dlt. In SPLIT~\cite{split}, one-step errors collected during driving train a local Gaussian process for each cell of a grid over the front and rear slip angles and the commanded torque. Carron et~al.~\cite{carron2019} extend an offline Gaussian-process model of a robotic arm by a disturbance estimate that is updated online. Xue et~al.~\cite{xue2024} regress local error dynamics on previous laps, with experiments on an IAC car at Putnam.

In comparison, \dlt is simpler and just stores a mean in a cell, which slowly forgets old data. The same layer therefore works with hosts whether or not they use a network. We also examine a change of track, not repeated laps.

\section{Vehicle Model and Prediction Task}
\label{sec:setting}

\subsection{Single-Track Host Model}
\begin{figure}[t]
\centering
\begin{tikzpicture}[font=\footnotesize, >=Latex, scale=0.9]
  \draw[gray, thick] (0,0) -- (3.0,0);
  \fill (1.33,0) circle (1.6pt);
  \node[anchor=south west, inner sep=1pt] at (1.38,0.04) {CG};
  \draw[fill=gray!25] (-0.3,-0.09) rectangle (0.3,0.09);
  \begin{scope}[shift={(3.0,0)}, rotate=18]
    \draw[fill=gray!25] (-0.3,-0.09) rectangle (0.3,0.09);
    \draw[->, cE, thick] (0,0.12) -- (0,0.75) node[above] {$F_{y,f}$};
    \draw[dashed, gray] (-0.5,0) -- (0.6,0);
  \end{scope}
  \draw[gray] (3.0,0) -- (3.6,0);
  \draw[gray] (3.45,0) arc (0:18:0.45) node[midway, right=-1pt] {$\delta$};
  \draw[->, cE, thick] (0,0.12) -- (0,0.75) node[above] {$F_{y,r}$};
  \draw[->, cDDM, thick] (0.32,0) -- (0.9,0) node[below, pos=0.55] {$F_x$};
  \draw[->, thick] (1.33,0) -- (2.2,0) node[above, pos=0.9] {$v_x$};
  \draw[->, thick] (1.33,0) -- (1.33,0.6) node[right] {$v_y$};
  \draw[->, thick] (1.33,0) ++(-150:0.38) arc (-150:-40:0.38) node[right] {$r$};
  \draw[|<->|, gray] (0,-0.66) -- (1.33,-0.66) node[midway, below] {$l_r$};
  \draw[|<->|, gray] (1.33,-0.66) -- (3.0,-0.66) node[midway, below] {$l_f$};
  \begin{scope}[shift={(4.9,-0.62)}]
    \draw[thick] (0,0) -- ({2.0*cos(15)},{2.0*sin(15)});
    \draw[gray] (0,0) -- (2.0,0);
    \draw[gray] (1.1,0) arc (0:15:1.1) node[midway, right] {$\theta$};
    \begin{scope}[shift={({1.0*cos(15)},{1.0*sin(15)})}, rotate=15]
      \draw[fill=gray!25] (-0.35,0) rectangle (0.35,0.22);
      \fill (0,0.11) circle (1.4pt);
    \end{scope}
    \draw[->, thick] ({1.0*cos(15)-0.03},{1.0*sin(15)+0.15}) -- ++(0,-0.75)
      node[below] {$mg$};
    \draw[->, cE, thick] ({1.0*cos(15)-0.02},{1.0*sin(15)+0.24})
      -- ++({-0.55*sin(15)},{0.55*cos(15)}) node[above] {$F_z$};
  \end{scope}
\end{tikzpicture}
\caption{The physics host: single-track model (left) and bank angle $\theta$ (right).}
\label{fig:host}
\end{figure}
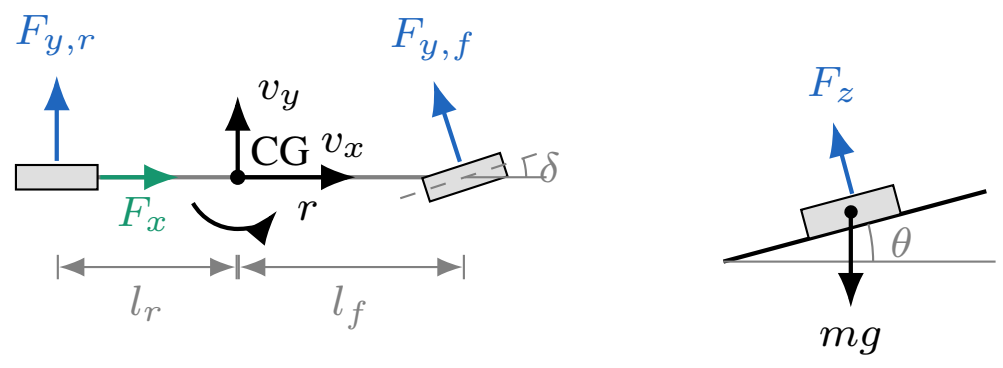

\begin{table}[t]
\caption{Parameters of the physics host: fixed (top) and fitted on Putnam (bottom).}
\label{tab:host}
\centering\scriptsize
\setlength{\tabcolsep}{3pt}
\begin{tabular}{@{}lrl@{\hspace{12pt}}lrl@{}}
\toprule
$m$ & 803 & kg & $C$ & 1.3 & -- \\
$l_f$ / $l_r$ & 1.657 / 1.315 & m & $p_K$ / $p_D$ & 2.0 / $-0.1$ & -- \\
$\rho$ & 1.226 & kg/m$^3$ & & & \\
\midrule
$I_z$ & 868 & kg\,m$^2$ & $\mu_f$ / $\mu_r$ & 1.70 / 1.52 & -- \\
$C_{m1}$ & 6538 & N & $k_f$ / $k_r$ & 11.5 / 14.1 & 1/rad \\
$C_{m2}$ & 8.42 & N\,s/m & $C_{L}A$ & 3.69 & m$^2$ \\
$C_{r0}$ & 463 & N & $b$ & 0.418 & -- \\
$C_d$ & 0.955 & kg/m & $G_f$ & $-7.9\cdot 10^{-4}$ & rad \\
$C_B$ & 2.68 & N/kPa & & & \\
\bottomrule
\end{tabular}
\end{table}

\dlt can be attached to any model that predicts the next state, including $v_x$, $v_y$ and $r$, and can continue from a corrected state. As our own host we use the single-track model of Fig.~\ref{fig:host}. With the slip angles $\alpha_f = \delta - \arctan\frac{v_y + l_f r}{v_x} + G_f$ and $\alpha_r = \arctan\frac{l_r r - v_y}{v_x}$, the lateral force of each axle $i \in \{f, r\}$ follows a Pacejka-type formula~\cite{pacejka2006},
\begin{align}
  F_{y,i} &= D_i \sin\!\bigl(C \arctan\tfrac{K_i \alpha_i}{C D_i}\bigr),
  \nonumber\\
  D_i &= \mu_i\bigl(1 + p_{D}\,(F_{z,i}/F_{z0,i} - 1)\bigr) F_{z,i},
  \nonumber\\
  K_i &= k_i F_{z0,i}\, \frac{\sin(2\arctan(F_{z,i}/(p_{K}F_{z0,i})))}
        {\sin(2\arctan(1/p_{K}))}.
  \label{eq:tire}
\end{align}
The normal loads $F_{z,i} = F_{z0,i}(\cos\theta + v_x r \sin\theta/g) + b_i\,\tfrac12\rho\, C_{L}A\, v_x^2$ consist of the static loads $F_{z0,i}$, the influence of the bank angle and the downforce, with $b_f = b$ and $b_r = 1 - b$ ($\tan\theta$, which matches $\dot\psi$ in Eq.~\eqref{eq:kin}, changes the FDE by at most 0.02\pc). The longitudinal force is $F_x = (C_{m1} - C_{m2} v_x)\,u_\mathrm{thr} - C_B\, u_\mathrm{br} - C_{r0} - C_d v_x^2$. To account for combined slip, the rear lateral force is multiplied by $\sqrt{1 - \min(1, (F_x/D_r)^2)}$. The state derivative $\mathbf F(\s,\uu)$ is
\begin{align}
  \dot v_x &= (F_x - F_{y,f}\sin\delta)/m + v_y r, \nonumber\\
  \dot v_y &= (F_{y,r} + F_{y,f}\cos\delta)/m - v_x r + g\sin\theta,
  \nonumber\\
  \dot r &= (l_f F_{y,f}\cos\delta - l_r F_{y,r})/I_z, \nonumber\\
  \dot\psi &= \frac{r}{\cos\theta},\quad
  \dot x = v_x \cos\psi - v_y \cos\theta \sin\psi, \nonumber\\
  \dot y &= v_x \sin\psi + v_y \cos\theta \cos\psi .
\label{eq:kin}
\end{align}
Slip angles, drive model and equations of motion follow~\cite{chrosniak2024}, without the rear offset $G_r$. The brake term, the bank-angle terms, the normal loads and the combined-slip factor are our additions. Mass and geometry for IAC cars are taken from~\cite{yahya2023} (44\pc static front-axle load; 58\pc in the DDM configuration, also used by \dyn), $C$ from~\cite{pacejka2006}; $p_K$ and $p_D$ are set by hand. The remaining parameters (13, Table~\ref{tab:host}) were fitted on the source track, minimizing the position error of 50-step rollouts. Their initial values come from physical limits and~\cite{froemmig2023}, those of $k_f$ and $k_r$ from~\cite{chrosniak2024}, which also gives the bounds of $I_z$, $C_d$ and $G_f$. The initial value of $I_z$ and one of the starts of $k_f$ and $k_r$ were taken from an earlier host fitted on LVMS A, i.e., on test data. The start with the lowest loss on the source track was kept. With neutral initial values, this loss rises by 21\pc, and the FDE of the physics model with \dlt by 0.6\pc at 0.6\,s and 11\pc at 2\,s in the DDM protocol. The fitted parameters are effective values and likely absorb neglected effects such as lateral load transfer. The only track information of the host is the bank angle, measured at the start of each horizon and held constant. On Putnam, $\theta = 0$, as it is a flat track. On BETTY, $\theta$ is estimated from the IMU.

\subsection{Prediction Task}
The state $\s = (x, y, \psi, v_x, v_y, r)$ holds position, heading, body-frame velocities and yaw rate, and the input $\uu$ holds the steering angle $\delta$, the throttle $u_\mathrm{thr} \in [0, 1]$ and the brake pressure $u_\mathrm{br}$ in kPa. Given the state $\s_k$ at time $k$ and the inputs of the next $K$ steps, a predictor returns the states along this horizon. The sampling rate is 25\,Hz ($\Delta t = 0.04$\,s).

We evaluate the mean position error over all starts for two horizons, $K = 15$ steps (0.6\,s) and $K = 50$ steps (2\,s), and once after a single step. At these two horizons, we call this the final displacement error (FDE). \emph{Zero-shot} describes a model that has not seen data of the target track. With \dlt, a host uses measurements of the target track while driving and is no longer zero-shot. As references, we use two predictors without a dynamics model. \emph{Extrapolation} holds $v_x$, $v_y$ and $r$ at their start values. The simple \emph{oracle} integrates the measured future $v_x$, $v_y$ and $r$ with the kinematics of Eq.~\eqref{eq:kin} and the measured bank angles, showing how much of the error a dynamics model could remove. It is not a strict bound, though: a model fitted on the positions of the test track can absorb part of the channel errors (Table~\ref{tab:intrack}).

\section{DELTA}
\label{sec:method}

\begin{figure}[t]
\centering
\begin{tikzpicture}[font=\footnotesize, >=Latex,
  box/.style={draw, rounded corners=2pt, align=center, minimum height=0.8cm,
              inner sep=2.5pt, fill=white},
  lab/.style={font=\scriptsize\itshape, text=gray}]
\node[box] (host) {host $\Phi$\\one step};
\node[box, right=0.55cm of host] (res) {residual $\e_k$\\Eq.~\eqref{eq:res}};
\node[box, right=0.55cm of res] (tab) {table $\dd_c$, Eq.~\eqref{eq:table}\\fast term $\bar\e_k$};
\draw[->] (host) -- node[above, font=\scriptsize]{$\hat\vv_k$} (res);
\draw[->] (res) -- (tab);
\node[above=0.35cm of res, font=\scriptsize] (meas) {measured $\vv_k$};
\draw[->] (meas) -- (res);
\node[lab, anchor=south west] at ([yshift=0.62cm]host.north west) {once per measurement};
\node[box, below=0.6cm of host] (step) {host $\Phi$\\$\shat_t \to \shat_{t+1}$};
\node[box, right=0.55cm of step] (plus) {$+\,\Delta t\,\mathbf B\,\dd_t$\\Eq.~\eqref{eq:roll}};
\node[box, below=0.6cm of tab] (corr) {correction $\dd_t$\\Eq.~\eqref{eq:dist}};
\draw[->] (step) -- (plus);
\draw[->] (corr) -- (plus);
\draw[->] (tab) -- node[right, font=\scriptsize]{look-up at $\shat_t$} (corr);
\draw[->] (plus.south) -- ++(0,-0.3) -| node[pos=0.25, below, font=\scriptsize]{next step} (step.south);
\node[lab, anchor=south west] at ([yshift=0.12cm]step.north west) {every step of every candidate rollout};
\end{tikzpicture}
\caption{\dlt around a host model: update (top) and rollout (bottom).}
\label{fig:arch}
\end{figure}
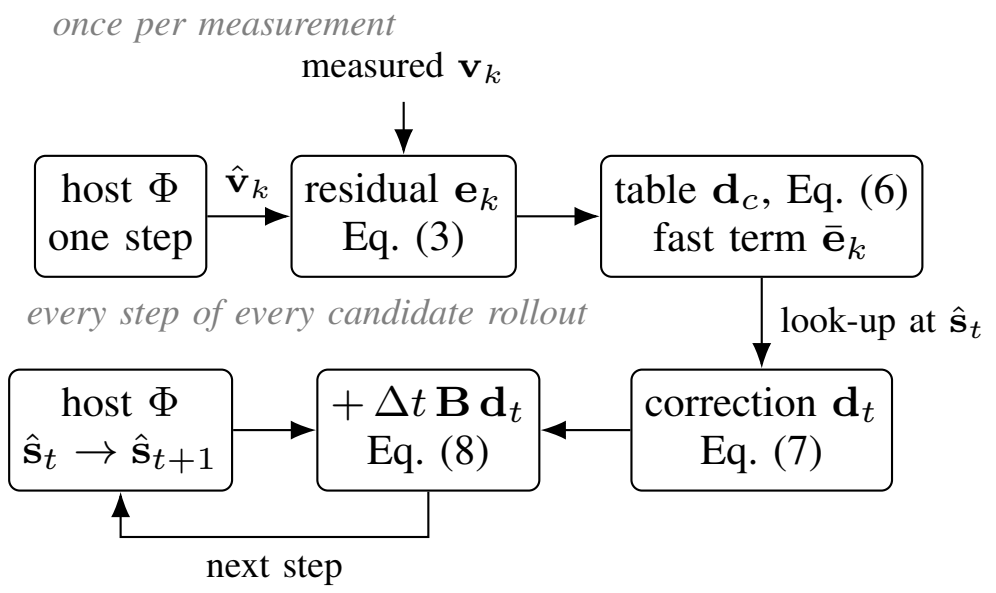

Whenever a new measurement arrives, \dlt compares it with what the host had predicted. This error is stored for the current driving situation and added to later predictions. Fig.~\ref{fig:arch} gives an overview and Fig.~\ref{fig:code} shows it as code.

\subsection{Residuals and Table}
We denote the one-step prediction of the host by $\Phi$, which is $\Phi(\s,\uu) = \s + \Delta t\,\mathbf F(\s,\uu)$ for the physics host. \dlt corrects $\vv = (v_x, v_y, r)$. With the next sample, \dlt forms the residual as in~\cite{kabzan2019}, divided by $\Delta t$,
\begin{equation}
  \e_k = \frac{\vv_k - \Phi_v(\s_{k-1}, \uu_{k-1})}{\Delta t}.
  \label{eq:res}
\end{equation}
Its unit is that of an acceleration. Besides the model error, it contains measurement errors, and \dlt corrects both alike. Non-finite residuals and those with $|e_{v_x}| \ge 30$\,m/s$^2$, $|e_{v_y}| \ge 30$\,m/s$^2$ or $|e_{r}| \ge 3$\,rad/s$^2$ are discarded.

The error of a host depends mainly on speed and cornering, so it is not constant. Each residual falls into one of 28 table cells, spanned by seven speed bands and four bands of steady-state lateral acceleration $|v_x r|$. Values beyond the limits fall into the outer bands. An exponentially weighted sum is kept for each cell $c$, and one for all residuals:
\begin{align}
  C_{s,c} &\leftarrow \lambda_c C_{s,c} + \e_k, &
  C_{n,c} &\leftarrow \lambda_c C_{n,c} + 1, \\
  G_s &\leftarrow \lambda_g G_s + \e_k, &
  G_n &\leftarrow \lambda_g G_n + 1.
\end{align}
Each cell mean is then pulled towards the global mean,
\begin{equation}
  \dd_c = \frac{C_{s,c} + \kappa\, G_s/G_n}{C_{n,c} + \kappa},
  \label{eq:table}
\end{equation}
so that a cell with few residuals starts close to the global estimate. The table is used once 500 valid residuals have been collected, i.e., after about 20\,s of driving.

\subsection{Correction in the Rollout}
The table updates slowly due to the weighted sums. Especially when the driving situation changes quickly, as at the entry of a corner, the table has not caught up yet but the new error is already visible in the latest residuals. \dlt therefore keeps the mean $\bar\e_k$ over the five latest residuals as a fast term, whose difference to the table value decays over the horizon. Unlike the table, it acts from the first residual on. Only residuals at most six samples old count, so after a gap in the data the fast term is zero until new residuals arrive. For a rollout from time $k$, the correction in step $t$ is
\begin{equation}
  \dd_t = \dd_{c(\shat_t)}
        + \bigl(\bar\e_k - \dd_{c(\s_k)}\bigr)\, e^{-t\Delta t/\tau},
  \label{eq:dist}
\end{equation}
and it enters the rollout as
\begin{equation}
  \shat_{t+1} = \Phi(\shat_t, \uu_t) + \Delta t\, \mathbf B\, \dd_t .
  \label{eq:roll}
\end{equation}
Through the matrix $\mathbf B = [\mathbf 0_{3\times 3}\ \ \mathbf I_3]^\top$, $\dd_t$ acts on $v_x$, $v_y$ and $r$, while position and heading follow from these velocities. Since the table is read at the predicted state $\shat_t$, the correction changes when the prediction changes cells. For \dyn, the network's history is continued with the corrected states.

\begin{figure}[t]
\begin{lstlisting}
import numpy as np
SPEED, LAT = [10, 20, 30, 40, 50, 60], [2, 6, 12]
LAM_G, LAM_C, KAPPA, WARM, TAU = 1-1/2000, 1-1/1000, 50, 500, .3
MAX_E = np.array([30.0, 30.0, 3.0])   # outlier limits
def cell(vx, r):                      # 7 x 4 = 28 cells
    return (np.searchsorted(SPEED, vx, 'right') * 4
            + np.searchsorted(LAT, abs(vx * r), 'right'))
class Delta:
    def __init__(self):
        self.Gs, self.Gn = np.zeros(3), 0.0
        self.Cs, self.Cn = np.zeros((28, 3)), np.zeros(28)
        self.recent, self.n = [], 0
    def update(self, v_meas, v_pred, vx, r, dt):  # vx, r: start
        e = (v_meas - v_pred) / dt    # residual, Eq. (3)
        if not np.all(np.abs(e) < MAX_E): return
        c, self.n = cell(vx, r), self.n + 1
        self.Gs, self.Gn = LAM_G*self.Gs + e, LAM_G*self.Gn + 1
        self.Cs[c] = LAM_C * self.Cs[c] + e
        self.Cn[c] = LAM_C * self.Cn[c] + 1
        self.recent = (self.recent + [e])[-5:]
    def table(self, vx, r):           # Eq. (6)
        if self.n < WARM: return np.zeros(3)
        c, g = cell(vx, r), self.Gs / self.Gn
        return (self.Cs[c] + KAPPA*g) / (self.Cn[c] + KAPPA)
    def correction(self, t, dt, s0, s):   # Eq. (7)
        fast = (np.mean(self.recent, 0) - self.table(s0.vx, s0.r)
                if self.recent else 0.0)
        return self.table(s.vx, s.r) + fast * np.exp(-t*dt/TAU)
def rollout(host, delta, s0, inputs, dt):  # Eq. (8)
    s, path = s0, []
    for t, u in enumerate(inputs):
        d = delta.correction(t, dt, s0, s)
        s = host(s, u)
        s.v = s.v + dt * d            # corrects vx, vy, r
        path.append(s.xy)
    return path
\end{lstlisting}
\caption{Simplified Python code of \dlt (no age limit of the fast term).}
\label{fig:code}
\end{figure}

One update per measurement takes 0.005\,ms. Within a control cycle, the table is the same for all candidates and steps, but each candidate reads it at its predicted states. A controller thus treats a host with \dlt like one without.

\begin{table*}[t]
\caption{RACECAR, Putnam $\to$ LVMS: FDE in m and change vs.\ extrapolation. DynSSM protocol: \dyn without bank angle.}
\label{tab:main}
\centering\scriptsize
\setlength{\tabcolsep}{7pt}
\begin{tabular}{@{}llcccc@{}}
\toprule
 & & \multicolumn{2}{c}{DDM protocol} &
   \multicolumn{2}{c}{DynSSM protocol} \\
\cmidrule(lr){3-4}\cmidrule(lr){5-6}
 & Predictor & 0.6\,s & 2\,s & 0.6\,s & 2\,s \\
\midrule
References & Extrapolation & 0.178 & 0.971 & 0.230 & 1.446 \\
 & Oracle & 0.157 ($-$12\pc) & 0.504 ($-$48\pc) & 0.177 ($-$23\pc) & 0.573 ($-$60\pc) \\
\addlinespace
Zero-shot & Physics & 0.196 (+10\pc) & 1.469 (+51\pc) & 0.283 (+23\pc) & 2.329 (+61\pc) \\
 & \ddmh{15}~\cite{chrosniak2024} & 0.311 (+75\pc) & 3.348 (+245\pc) & -- & -- \\
 & \ddmh{50} & 0.223 (+25\pc) & 1.691 (+74\pc) & -- & -- \\
 & \dyn & 0.188 (+6\pc) & 1.745 (+80\pc) & 0.236 (+3\pc) & 2.816 (+95\pc) \\
\addlinespace
With \dlt & Extrapolation & 0.173 ($-$3\pc) & 1.008 (+4\pc) & 0.206 ($-$10\pc) & 1.460 (+1\pc) \\
 & Physics & 0.161 ($-$10\pc) & 0.584 ($-$40\pc) & 0.194 ($-$15\pc) & 0.860 ($-$41\pc) \\
 & \ddmh{15} & 0.195 (+9\pc) & 1.507 (+55\pc) & -- & -- \\
 & \ddmh{50} & 0.176 ($-$1\pc) & 1.115 (+15\pc) & -- & -- \\
 & \dyn & 0.160 ($-$10\pc) & 0.602 ($-$38\pc) & 0.194 ($-$16\pc) & 1.128 ($-$22\pc) \\
\bottomrule
\end{tabular}
\end{table*}

\section{Experimental Setup}
\label{sec:setup}

\subsection{Datasets}
In the main experiments, we test \dlt on recordings of an IAC car that are distributed with the DDM repository and attributed to the RACECAR dataset~\cite{racecar} in~\cite{chrosniak2024,dynssm}. We refer to them as RACECAR. For a direct comparison with DDM and DynSSM, we copy their protocols: in the DDM protocol, parts of recording A of LVMS are the test data, in the DynSSM protocol, A and B are the test data. With the scripts of the DDM repository, we could not exactly reproduce the sample counts reported in~\cite{chrosniak2024} (13\,418 for training, more than run4\_2 holds), while our DynSSM protocol matches the 56\,602 samples of~\cite{dynssm} (rows of A and B; 56\,544 starts; for 2\,s, the last row is repeated). The source for the physics host and DDM is the Putnam Park recording run4\_2 (8\,min). We use a third protocol for ablations only: LVMS recordings A and B processed with the DDM parser. The parsed sections reach velocities of up to 36\,m/s, while recording A also has a section with up to 54\,m/s, which we report separately. Results on RACECAR, evaluated repeatedly during development, are descriptive; only the BETTY hypotheses were fixed beforehand.

We also test the models on a second dataset, BETTY~\cite{betty}. Its recordings come from an \mbox{AV-21} of another IAC team. As the car differs, we fitted the hosts on Monza data, the physics host from neutral initial values, and tested on two LVMS recordings. Like Putnam Park, Monza is a road course.

\subsection{Host Models}
DDM runs with the code of its repository, which selects a subset of run4\_2 for training, and with its evaluation script, which holds the estimated coefficients over the horizon. We changed only the evaluated range (the whole parsed recording instead of the first, much easier 3\,000 samples, with the unchanged script: 0.324 instead of 0.142\,m FDE at 0.6\,s with history 15 and without bank angle; \cite{chrosniak2024} reports 0.384\,m for its test set of 10\,606 samples), the bank-angle kinematics (the FDE is 0.5 to 7\pc higher without it), the alignment of start pose and start velocities, which the script places 14 (history 15) or 49 samples (history 50) apart (aligned, the 2\,s FDE changes by +1 to +10\pc and $-$9 to $-$18\pc), and the call of \dlt. We report \ddmh{15} with the original history of 15 samples and \ddmh{50}, retrained with 50 so that the estimation window spans the 2\,s of held coefficients, one seed each; the star marks our version, as for \dyn. As the script reports only aggregated errors, \ddms runs on its own starts (13\,319 to 13\,389 instead of 13\,356), without intervals, only under the DDM protocol and not in Fig.~\ref{fig:horizon}. \dyn is a re-implementation of DynSSM~\cite{dynssm}, rolled out recursively like the original and trained on run2\_1 and twice on run4\_2, the only combination matching the 62\,419 samples of three laps in its paper (rows; 56\,554 training windows). The result of 0.236\,m FDE at 0.6\,s under the DynSSM protocol is close to the reported FDE of 0.246\,m~\cite{dynssm}. A few design choices of \dyn were made with LVMS results in view, which can only favor it. \dyn was trained with five seeds, and we report the mean unless stated otherwise; at 2\,s, single seeds alone reach 1.43 to 2.10\,m in the DDM protocol and 1.89 to 4.72\,m in the DynSSM protocol. Choices that~\cite{dynssm} leaves open, such as layer sizes and history, are listed in the repository. In the DDM protocol, both networks integrate their predicted velocities with $\dot\psi$, $\dot x$ and $\dot y$ from Eq.~\eqref{eq:kin}, using the bank angle measured at the start of the horizon, as the physics host does. \dyn does the same on BETTY. Their dynamics contain no bank term, so their residuals also contain the missing term $g\sin\theta$. Adding it to \dyn, which needs no retraining as $\theta = 0$ on Putnam, lowers the zero-shot FDE at 2\,s from 1.745 to 1.259\,m, but raises it with \dlt from 0.602 to 0.756\,m, for which we had not yet found the cause. In the DynSSM protocol, it integrates on a flat plane, which we assume matches the original~\cite{dynssm}. P is the physics host, and $^+$ marks \dlt.

\dlt always starts with an empty table and with the hyperparameters of Fig.~\ref{fig:code}. These were chosen on recordings outside the test sets (including another BETTY recording of LVMS). Halving or doubling one of the hyperparameters changed the 2\,s FDE by 8\pc at most. Shifting the speed bands by 5\,m/s raised it by up to 9.7\pc on LVMS A. Within a recording, \dlt is never reset; the table continues across gaps in the data, but no residual spans a gap. Starts with missing channels in their window are not evaluated.

Consecutive starts share most of their horizon. Intervals therefore come from a paired bootstrap over blocks of 30\,s (2\,000 resamples), at the 95\pc level unless stated otherwise. They cover the variation over time within the recordings, not the spread over the five training seeds of the networks.

\section{Results}
\label{sec:results}

\subsection{Open-Loop Prediction}

With no target-track data available (zero-shot), extrapolation shows a lower FDE than each of the three host models at 2\,s (Table~\ref{tab:main}). Under the DDM protocol, the physics model shows an FDE 51\pc above, \ddmh{15} 245\pc above (\ddmh{50}: 74\pc) and the simple oracle an FDE 48\pc below extrapolation. This leaves room for improvement.

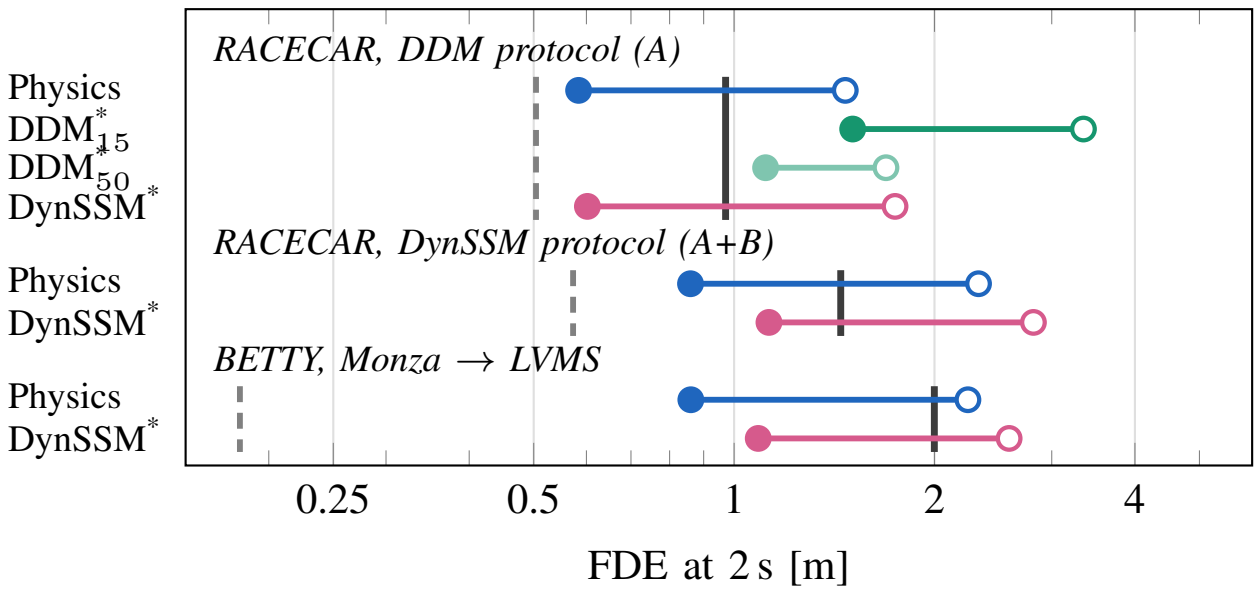
\begin{figure}[t]
\centering
\begin{tikzpicture}
\begin{axis}[width=\columnwidth, height=4.6cm,
  xmode=log, xmin=0.15, xmax=6, ymin=-0.1, ymax=11.7, y dir=reverse,
  xtick={0.25,0.5,1,2,4}, xticklabels={0.25,0.5,1,2,4},
  minor xtick={0.2,0.3,0.4,0.6,0.7,0.8,0.9,3,5},
  ytick={2,3,4,5,7,8,10,11},
  yticklabels={Physics,\ddmh{15},\ddmh{50},DynSSM\textsuperscript{*},Physics,DynSSM\textsuperscript{*},Physics,DynSSM\textsuperscript{*}},
  xlabel={FDE at 2\,s [m]},
  tick label style={font=\footnotesize}, yticklabel style={font=\scriptsize, text width=1.05cm, align=left}, label style={font=\footnotesize},
  ytick style={draw=none}, xmajorgrids, grid style={gray!25},
  clip=false]
\node[anchor=west, font=\scriptsize\itshape] at (axis cs:0.155,1) {RACECAR, DDM protocol (A)};
\node[anchor=west, font=\scriptsize\itshape] at (axis cs:0.155,6) {RACECAR, DynSSM protocol (A+B)};
\node[anchor=west, font=\scriptsize\itshape] at (axis cs:0.155,9) {BETTY, Monza $\to$ LVMS};
\foreach \xf/\xo/\ya/\yb in {0.971/0.504/1.65/5.35, 1.446/0.573/6.65/8.35,
                             1.999/0.181/9.65/11.35}{
  \edef\tmp{\noexpand\draw[cF, line width=1.3pt] (axis cs:\xf,\ya) -- (axis cs:\xf,\yb);
            \noexpand\draw[gray, line width=1.1pt, densely dashed] (axis cs:\xo,\ya) -- (axis cs:\xo,\yb);}
  \tmp
}
\foreach \col/\za/\wa/\y in {
  cE/1.469/0.584/2, cDDM/3.348/1.507/3, cDDM!55/1.691/1.115/4, cDyn/1.745/0.602/5,
  cE/2.329/0.860/7, cDyn/2.816/1.128/8,
  cE/2.245/0.861/10, cDyn/2.589/1.087/11}{
  \edef\tmp{\noexpand\draw[\col, line width=1.1pt] (axis cs:\za,\y) -- (axis cs:\wa,\y);
            \noexpand\node[circle, draw=\col, fill=white, inner sep=1.5pt, thick] at (axis cs:\za,\y) {};
            \noexpand\node[circle, draw=\col, fill=\col, inner sep=1.5pt, thick] at (axis cs:\wa,\y) {};}
  \tmp
}
\end{axis}
\end{tikzpicture}
\caption{FDE at 2\,s for the hosts alone (open) and with \dlt (filled); solid bars: extrapolation, dashed bars: oracle.}
\label{fig:dumbbell}
\end{figure}

Adding \dlt uses most of this room (Fig.~\ref{fig:dumbbell}): P$^+$ closes 83\pc of the gap between extrapolation and oracle at 2\,s (DDM protocol). At 2\,s, \dlt cuts the error of the physics model, \ddmh{15} and \dyn by 55 to 66\pc (\dyn per seed 56 to 72\pc; 48 to 69\pc in the DynSSM protocol), that of \ddmh{50} by 34\pc. At 2\,s, the physics model is ahead by 3\pc under the DDM protocol (interval $-$3.9 to 8.5\pc; per seed of \dyn $-$4 to 9\pc). Under the DynSSM protocol, it is 23\pc ahead of \dynp with bank angle (1.112\,m; interval 14 to 30\pc; per seed $-$3 to 44\pc). With neutral initial values (Sec.~\ref{sec:setting}), however, P$^+$ is 8\pc behind \dynp under the DDM protocol (interval 1 to 16\pc), so the ranking depends on the fit. \ddmp{15} reaches 1.507\,m and \ddmp{50} 1.115\,m, 55 and 15\pc above extrapolation and 2.6 and 1.9 times P$^+$. With extrapolation as host, \dlt changes the FDE by only $-$10 to +4\pc, so its gain at 2\,s needs a dynamics model.

\begin{table}[t]
\caption{BETTY, Monza $\to$ LVMS: FDE at 2\,s with 99\pc intervals.}
\label{tab:betty}
\centering\scriptsize
\setlength{\tabcolsep}{2.5pt}
\begin{tabular}{@{}llcc@{}}
\toprule
 & $a$ vs.\ $b$ & FDE $a$ / $b$ [m] & $a/b - 1$ \\
\midrule
H1 & P$^+$ vs.\ extrapolation & 0.861 / 1.999 & $-$56.9\pc [$-$66.5, $-$44.2] \\
H2 & P$^+$ vs.\ P & 0.861 / 2.245 & $-$61.7\pc [$-$68.7, $-$50.5] \\
H3 & P$^+$ vs.\ P$^+$, offline & 0.861 / 1.667 & $-$48.4\pc [$-$56.4, $-$36.4] \\
H4 & DynSSM\textsuperscript{*+} vs.\ \dyn & 1.087 / 2.589 & $-$58.0\pc [$-$63.5, $-$49.7] \\
H5 & P$^+$ vs.\ DynSSM\textsuperscript{*+}, no bank & 0.924 / 1.141 & $-$19.1\pc [$-$36.3, +2.0] \\
\bottomrule
\end{tabular}
\end{table}

\emph{Second vehicle.} On BETTY, we fixed five hypotheses before the evaluation and test them at 2\,s (Table~\ref{tab:betty}). H1 to H4 test for a difference, H5 for non-inferiority with a 5\pc margin. The data support all five. H5 compares the physics model and \dyn, both with \dlt and without the bank angle. The physics model is at most 5\pc worse than \dyn, but we cannot conclude that it is better. With \dlt, both hosts show almost the same median error, while the physics model is ahead mainly in the error tail. The BETTY dataset confirms the layer, but a ranking of the hosts needs more data or tests.

\begin{figure}[t]
\centering
\begin{tikzpicture}
\begin{axis}[xbar, scale only axis, width=4.9cm, height=3.1cm,
  bar width=2.7pt, y dir=reverse,
  xmin=-15, xmax=200, ymin=0.4, ymax=10.6,
  ytick={1,...,10},
  yticklabels={no yaw-rate correction, no $v_y$ correction,
    fast term only, lat.-acc.\ bands only, one cell (offset-free),
    no $v_x$ correction, table only, speed bands only,
    look-up at start state, no pooling ($\kappa = 0$)},
  xtick={0,50,100,150,200},
  xlabel={change of FDE at 2\,s [\%]},
  tick label style={font=\footnotesize}, label style={font=\footnotesize},
  ytick style={draw=none}, xmajorgrids, grid style={gray!25},
  legend style={font=\footnotesize, at={(0.97,0.60)}, anchor=east,
    draw=none, fill=white, fill opacity=0.8, text opacity=1},
  legend cell align=left, legend image code/.code={
    \draw[#1] (0cm,-0.08cm) rectangle (0.25cm,0.08cm);}]
\addplot[fill=cE, draw=cE] coordinates
  {(117.8,1) (50.7,2) (70.0,3) (6.2,4) (10.4,5) (32.5,6) (6.1,7) (6.1,8)
   (1.7,9) (-1.0,10)};
\addlegendentry{RACECAR (A+B)}
\addplot[fill=cE!35, draw=cE] coordinates
  {(183.9,1) (183.3,2) (91.9,3) (45.7,4) (40.0,5) (30.4,6) (21.4,7) (7.6,8)
   (3.3,9) (-7.3,10)};
\addlegendentry{BETTY}
\draw[black] (axis cs:0,0.4) -- (axis cs:0,10.6);
\end{axis}
\end{tikzpicture}
\caption{Ablations of \dlt on the physics model.}
\label{fig:abl}
\end{figure}
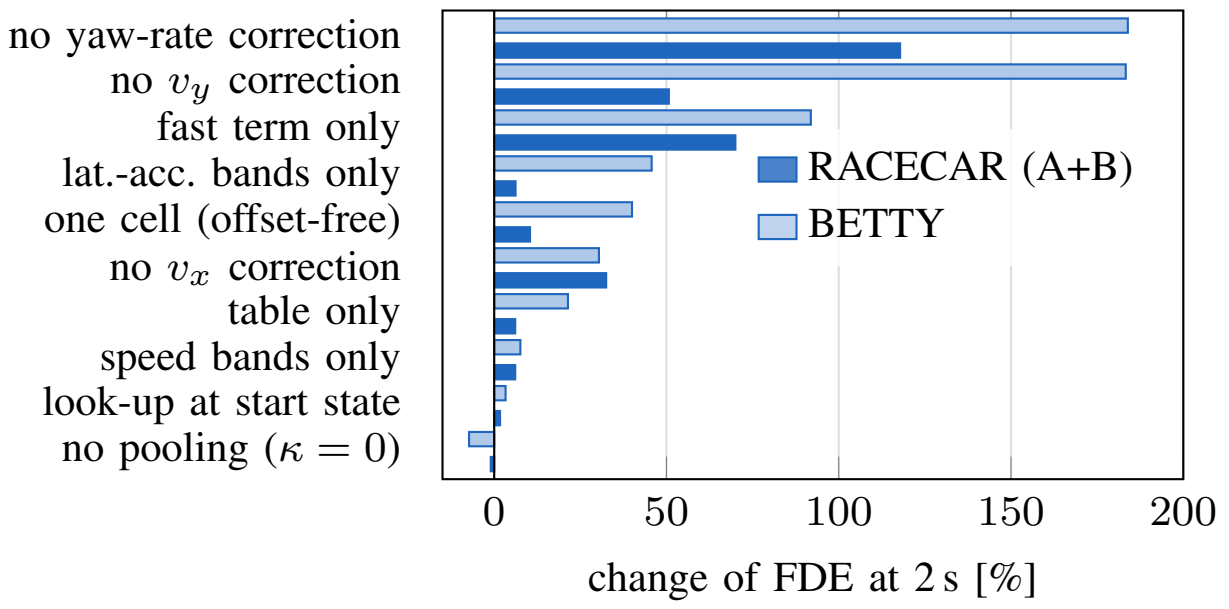

\emph{Ablations.} We removed or simplified one part of the layer at a time (Fig.~\ref{fig:abl}). Dropping the yaw-rate correction costs the most, on BETTY together with that of the lateral velocity. The table contributes more than the fast term. Using only one cell, as in classical offset-free estimation~\cite{pannocchia2003,maeder2009}, but with the fast term kept, raises the error by 10 to 40\pc.

A more predictable alternative is a table filled on the source track and kept fixed, while the fast term stays online. This helps as well, but it recovers no more than 49\pc of the online gain at 2\,s. On BETTY, \dlt is 48\pc better (H3). A table frozen at the end of A and reused from the start is 21\pc worse; the error changes over the drive.

With the tested data, the layer removes mostly lateral bias. On LVMS, the physics model predicts on average 0.75\,m to one side after 2\,s without \dlt and 0.09\,m with it. The layer also learns most of the contribution of the bank-angle term. Without this term, the physics model alone is 34 to 96\pc worse in the four evaluations, with \dlt only 7 to 23\pc. If both hosts run without the bank angle, P$^+$ is 7 to 8\pc behind \dynp with the DDM parser, 11\pc ahead in the DynSSM protocol and 19\pc on BETTY (H5).

\emph{Measurement effects.} Some of the position errors do not come from dynamics modeling. The position signal lags 33 to 41\,ms behind the velocities in the RACECAR LVMS recordings. This lag can be exploited with a position offset learned from past end-point errors. Such an offset reduces the 0.6\,s FDE by 28\pc for the physics model using \dlt and by 32\pc for the oracle. On BETTY, it gains less than 2\pc. No dynamics model fitted on another track can do so, and \dlt sees only velocity errors. It never goes below the oracle in any cross-track evaluation, and the simulation with exact signals points the same way (Sec.~\ref{sec:sim}).

\begin{table}[t]
\caption{Physics model fitted and tested on the same track. FDE in m at 0.6\,s / 2\,s.}
\label{tab:intrack}
\centering\scriptsize
\setlength{\tabcolsep}{2.5pt}
\begin{tabular}{@{}lcccc@{}}
\toprule
Fit $\to$ test & Extrapolation & Oracle & Physics & Physics + \dlt \\
\midrule
LVMS A $\to$ B & 0.179 / 1.268 & 0.105 / 0.313 & 0.188 / 0.970 & 0.128 / 0.781 \\
LVMS B $\to$ A & 0.178 / 0.971 & 0.157 / 0.504 & 0.150 / 0.775 & 0.162 / 0.529 \\
Putnam 2\_1 $\to$ 4\_2 & 0.220 / 1.872 & 0.151 / 0.580 & 0.120 / 0.769 & 0.134 / 0.677 \\
Putnam 4\_2 $\to$ 2\_1 & 0.189 / 1.152 & 0.175 / 0.627 & 0.068 / 0.272 & 0.136 / 0.397 \\
\bottomrule
\end{tabular}
\end{table}

\emph{Same-track evaluation.} When the host is fitted on the test track itself, the results differ. The physics model beats extrapolation at 2\,s in all pairs of Table~\ref{tab:intrack}, even without \dlt, which cuts the error by an additional 12 to 32\pc for three pairs. On A at 2\,s, P$^+$ fitted on Putnam (0.584\,m) even beats the physics model fitted on B (0.775\,m). At the same time, three of the four pairs show a worse FDE at 0.6\,s with \dlt. On Putnam, fitted on 4\_2 and tested on 2\_1, the physics model is already well below the oracle (0.068 against 0.175\,m) and \dlt doubles its FDE at 0.6\,s. This pair has only 2\,370 starts, about three blocks of 30\,s. In all eight combinations of pair and horizon, \dlt lowered the error where the host alone was above the oracle and raised it where it was below; without such a margin, it seems to add mainly the noise of its estimates.

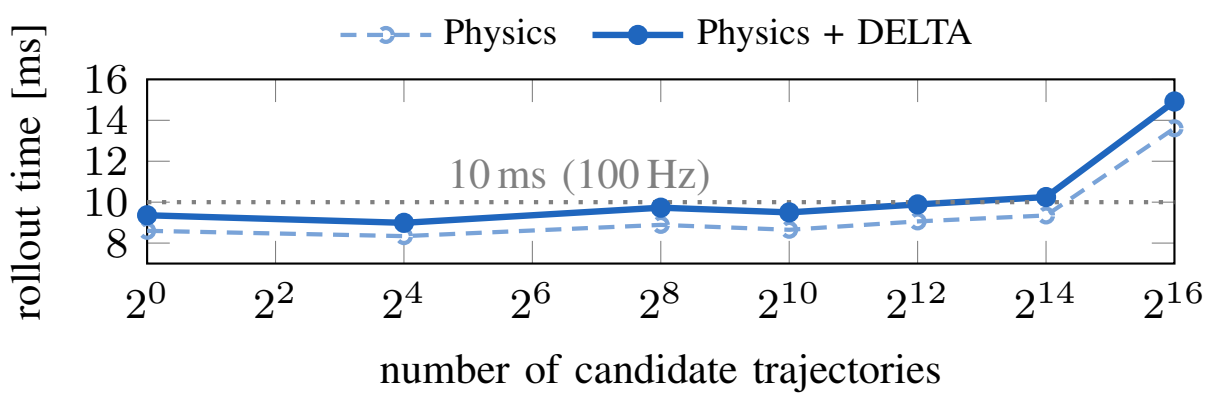
\begin{figure}[t]
\centering
\begin{tikzpicture}
\begin{semilogxaxis}[width=0.97\columnwidth, height=2.8cm,
  xlabel={number of candidate trajectories}, ylabel={rollout time [ms]},
  xmin=1, xmax=65536, ymin=7, ymax=16, ytick={8,10,12,14,16},
  log basis x=2, xtick={1,4,16,64,256,1024,4096,16384,65536},
  xticklabels={$2^{0}$,$2^{2}$,$2^{4}$,$2^{6}$,$2^{8}$,$2^{10}$,$2^{12}$,$2^{14}$,$2^{16}$},
  tick label style={font=\footnotesize}, label style={font=\footnotesize},
  legend columns=2, legend style={font=\scriptsize, at={(0.5,1.03)}, anchor=south,
  /tikz/every even column/.append style={column sep=4pt},
  draw=none, fill=none}, legend cell align=left]
\addplot[cE!60, thick, densely dashed, mark=o, mark size=1.3pt] coordinates
  {(1,8.60) (16,8.34) (256,8.89) (1024,8.65) (4096,9.06) (16384,9.35) (65536,13.61)};
\addlegendentry{Physics}
\addplot[cE, very thick, mark=*, mark size=1.3pt] coordinates
  {(1,9.36) (16,8.99) (256,9.73) (1024,9.50) (4096,9.89) (16384,10.25) (65536,14.92)};
\addlegendentry{Physics + \dlt}
\addplot[gray, dotted, thick, forget plot] coordinates {(1,10) (65536,10)};
\node[font=\footnotesize, gray, anchor=south west, inner sep=1pt] at (axis cs:24,10) {10\,ms (100\,Hz)};
\end{semilogxaxis}
\end{tikzpicture}
\caption{Time for one 2\,s rollout (50 steps) of the physics model, alone and with \dlt.}
\label{fig:latency}
\end{figure}

\subsection{Computation Time}
We timed the physics model alone and with \dlt in PyTorch on an RTX~5060~Ti (16\,GB) with CUDA graphs. A single 2\,s rollout takes 8.6\,ms alone and 9.4\,ms with \dlt (91.5\,ms without CUDA graphs). As the rollout consists of many small GPU operations, its time barely grows with the number of candidates (Fig.~\ref{fig:latency}): 16\,384 candidates take 9.3\,ms alone and 10.3\,ms with \dlt. Up to 16\,384 candidates, a rollout thus takes about 10\,ms, one cycle at about 100\,Hz, before the costs and the rest of the controller are added. For the 0.6\,s horizon of~\cite{chrosniak2024,dynssm}, a rollout with \dlt takes 2.8\,ms (3.0\,ms for 16\,384), and for the 43 steps of~\cite{dkmgp} 8.0\,ms (8.8\,ms). As orders of magnitude on other hardware, \cite{chrosniak2024} reports about 700\,Hz for one DDM inference (RTX~2080~Ti), \cite{dynssm} 2\,ms average inference time for DynSSM (RTX~4070~SUPER workstation), and \cite{dkmgp} 22 to 75\,Hz for 43 steps with DKMGP and 0.043\,Hz with DKL-SKIP~\cite{dklskip} (RTX~3080).

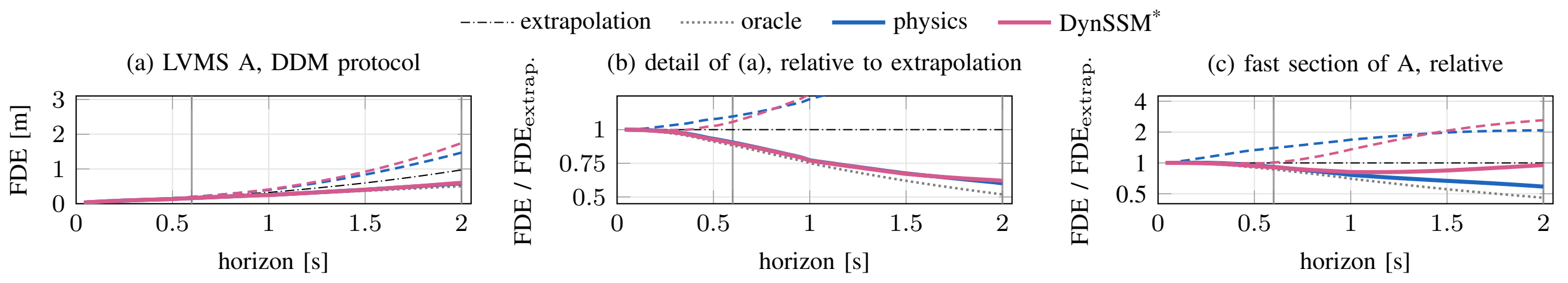
\begin{figure*}[t]
\centering
\begin{tikzpicture}
\tikzset{hzalone/.style={line width=0.8pt, dash pattern=on 3pt off 1.8pt},
  hzdelta/.style={line width=1.3pt}}
\begin{groupplot}[group style={group size=3 by 1, horizontal sep=1.65cm},
  width=0.34\textwidth, height=2.8cm, xmin=0, xmax=2.05,
  xtick={0,0.5,1,1.5,2}, xlabel={horizon [s]},
  tick label style={font=\footnotesize}, label style={font=\footnotesize},
  title style={font=\footnotesize, yshift=-3pt}, grid=major, grid style={gray!20},
  extra x ticks={0.6,2}, extra x tick labels={},
  extra x tick style={grid=major, major grid style={gray!75, line width=0.6pt}}]
\nextgroupplot[title={(a) LVMS A, DDM protocol}, ylabel={FDE [m]}, ymin=0, ymax=3.1,
  legend to name=hzlegend, legend columns=4,
  legend style={font=\footnotesize, draw=none, /tikz/every even column/.append style={column sep=7pt}}]
\addplot[black, thin, densely dashdotted] coordinates {(0.04,0.0326) (0.08,0.0462) (0.12,0.0577) (0.16,0.0698) (0.20,0.0772) (0.24,0.0895) (0.28,0.0970) (0.32,0.1057) (0.36,0.1152) (0.40,0.1219) (0.44,0.1310) (0.48,0.1389) (0.52,0.1523) (0.56,0.1655) (0.60,0.1782) (0.64,0.1926) (0.68,0.2056) (0.72,0.2199) (0.76,0.2344) (0.80,0.2482) (0.84,0.2640) (0.88,0.2789) (0.92,0.2950) (0.96,0.3107) (1.00,0.3224) (1.04,0.3428) (1.08,0.3631) (1.12,0.3836) (1.16,0.4041) (1.20,0.4247) (1.24,0.4469) (1.28,0.4686) (1.32,0.4910) (1.36,0.5138) (1.40,0.5366) (1.44,0.5610) (1.48,0.5846) (1.52,0.6103) (1.56,0.6368) (1.60,0.6634) (1.64,0.6915) (1.68,0.7196) (1.72,0.7486) (1.76,0.7785) (1.80,0.8087) (1.84,0.8403) (1.88,0.8721) (1.92,0.9049) (1.96,0.9382) (2.00,0.9709)};
\addplot[gray, thick, densely dotted] coordinates {(0.04,0.0326) (0.08,0.0461) (0.12,0.0574) (0.16,0.0691) (0.20,0.0760) (0.24,0.0877) (0.28,0.0944) (0.32,0.1020) (0.36,0.1101) (0.40,0.1151) (0.44,0.1220) (0.48,0.1273) (0.52,0.1381) (0.56,0.1481) (0.60,0.1574) (0.64,0.1679) (0.68,0.1764) (0.72,0.1860) (0.76,0.1950) (0.80,0.2032) (0.84,0.2129) (0.88,0.2209) (0.92,0.2302) (0.96,0.2387) (1.00,0.2423) (1.04,0.2533) (1.08,0.2641) (1.12,0.2748) (1.16,0.2851) (1.20,0.2950) (1.24,0.3059) (1.28,0.3157) (1.32,0.3259) (1.36,0.3360) (1.40,0.3455) (1.44,0.3559) (1.48,0.3651) (1.52,0.3757) (1.56,0.3863) (1.60,0.3967) (1.64,0.4077) (1.68,0.4182) (1.72,0.4289) (1.76,0.4398) (1.80,0.4503) (1.84,0.4614) (1.88,0.4721) (1.92,0.4830) (1.96,0.4939) (2.00,0.5040)};
\addplot[cE, hzalone, forget plot] coordinates {(0.04,0.0326) (0.08,0.0462) (0.12,0.0580) (0.16,0.0705) (0.20,0.0786) (0.24,0.0917) (0.28,0.1001) (0.32,0.1099) (0.36,0.1205) (0.40,0.1287) (0.44,0.1398) (0.48,0.1495) (0.52,0.1647) (0.56,0.1803) (0.60,0.1957) (0.64,0.2134) (0.68,0.2299) (0.72,0.2482) (0.76,0.2673) (0.80,0.2860) (0.84,0.3075) (0.88,0.3286) (0.92,0.3513) (0.96,0.3744) (1.00,0.3956) (1.04,0.4260) (1.08,0.4564) (1.12,0.4877) (1.16,0.5199) (1.20,0.5526) (1.24,0.5876) (1.28,0.6227) (1.32,0.6591) (1.36,0.6968) (1.40,0.7352) (1.44,0.7756) (1.48,0.8163) (1.52,0.8592) (1.56,0.9038) (1.60,0.9492) (1.64,0.9963) (1.68,1.0440) (1.72,1.0931) (1.76,1.1435) (1.80,1.1947) (1.84,1.2476) (1.88,1.3014) (1.92,1.3565) (1.96,1.4127) (2.00,1.4694)};
\addplot[cE, hzdelta] coordinates {(0.04,0.0326) (0.08,0.0461) (0.12,0.0576) (0.16,0.0695) (0.20,0.0766) (0.24,0.0886) (0.28,0.0956) (0.32,0.1034) (0.36,0.1118) (0.40,0.1171) (0.44,0.1243) (0.48,0.1301) (0.52,0.1412) (0.56,0.1516) (0.60,0.1613) (0.64,0.1722) (0.68,0.1812) (0.72,0.1912) (0.76,0.2008) (0.80,0.2095) (0.84,0.2196) (0.88,0.2281) (0.92,0.2377) (0.96,0.2461) (1.00,0.2491) (1.04,0.2619) (1.08,0.2744) (1.12,0.2870) (1.16,0.2990) (1.20,0.3106) (1.24,0.3238) (1.28,0.3357) (1.32,0.3480) (1.36,0.3603) (1.40,0.3720) (1.44,0.3849) (1.48,0.3962) (1.52,0.4099) (1.56,0.4239) (1.60,0.4377) (1.64,0.4525) (1.68,0.4665) (1.72,0.4811) (1.76,0.4959) (1.80,0.5103) (1.84,0.5256) (1.88,0.5405) (1.92,0.5558) (1.96,0.5709) (2.00,0.5841)};
\addplot[cDyn, hzalone, forget plot] coordinates {(0.04,0.0326) (0.08,0.0462) (0.12,0.0577) (0.16,0.0697) (0.20,0.0770) (0.24,0.0893) (0.28,0.0967) (0.32,0.1054) (0.36,0.1151) (0.40,0.1224) (0.44,0.1330) (0.48,0.1423) (0.52,0.1570) (0.56,0.1728) (0.60,0.1884) (0.64,0.2069) (0.68,0.2243) (0.72,0.2438) (0.76,0.2645) (0.80,0.2849) (0.84,0.3084) (0.88,0.3320) (0.92,0.3572) (0.96,0.3834) (1.00,0.4084) (1.04,0.4427) (1.08,0.4768) (1.12,0.5118) (1.16,0.5481) (1.20,0.5854) (1.24,0.6253) (1.28,0.6658) (1.32,0.7082) (1.36,0.7525) (1.40,0.7980) (1.44,0.8460) (1.48,0.8953) (1.52,0.9473) (1.56,1.0019) (1.60,1.0581) (1.64,1.1169) (1.68,1.1774) (1.72,1.2403) (1.76,1.3055) (1.80,1.3727) (1.84,1.4427) (1.88,1.5147) (1.92,1.5893) (1.96,1.6661) (2.00,1.7449)};
\addplot[cDyn, hzdelta] coordinates {(0.04,0.0326) (0.08,0.0461) (0.12,0.0576) (0.16,0.0694) (0.20,0.0765) (0.24,0.0883) (0.28,0.0953) (0.32,0.1031) (0.36,0.1113) (0.40,0.1165) (0.44,0.1236) (0.48,0.1294) (0.52,0.1404) (0.56,0.1508) (0.60,0.1605) (0.64,0.1714) (0.68,0.1805) (0.72,0.1905) (0.76,0.2001) (0.80,0.2089) (0.84,0.2190) (0.88,0.2277) (0.92,0.2373) (0.96,0.2455) (1.00,0.2482) (1.04,0.2610) (1.08,0.2735) (1.12,0.2860) (1.16,0.2980) (1.20,0.3095) (1.24,0.3226) (1.28,0.3343) (1.32,0.3465) (1.36,0.3588) (1.40,0.3703) (1.44,0.3831) (1.48,0.3944) (1.52,0.4082) (1.56,0.4227) (1.60,0.4369) (1.64,0.4525) (1.68,0.4675) (1.72,0.4832) (1.76,0.4995) (1.80,0.5156) (1.84,0.5330) (1.88,0.5502) (1.92,0.5680) (1.96,0.5858) (2.00,0.6022)};
\legend{extrapolation, oracle, physics, \dyn}
\nextgroupplot[title={(b) detail of (a), relative to extrapolation}, ylabel={FDE / FDE$_\mathrm{extrap.}$}, ymin=0.45, ymax=1.25, ytick={0.5,0.75,1}]
\addplot[black, thin, densely dashdotted, forget plot] coordinates {(0.04,1.0000) (2.00,1.0000)};
\addplot[gray, thick, densely dotted, forget plot] coordinates {(0.04,1.0000) (0.08,0.9977) (0.12,0.9944) (0.16,0.9908) (0.20,0.9853) (0.24,0.9803) (0.28,0.9729) (0.32,0.9649) (0.36,0.9554) (0.40,0.9442) (0.44,0.9310) (0.48,0.9163) (0.52,0.9069) (0.56,0.8949) (0.60,0.8832) (0.64,0.8714) (0.68,0.8579) (0.72,0.8459) (0.76,0.8322) (0.80,0.8187) (0.84,0.8064) (0.88,0.7922) (0.92,0.7804) (0.96,0.7680) (1.00,0.7515) (1.04,0.7390) (1.08,0.7274) (1.12,0.7164) (1.16,0.7054) (1.20,0.6946) (1.24,0.6845) (1.28,0.6738) (1.32,0.6638) (1.36,0.6538) (1.40,0.6438) (1.44,0.6345) (1.48,0.6245) (1.52,0.6155) (1.56,0.6067) (1.60,0.5980) (1.64,0.5896) (1.68,0.5811) (1.72,0.5729) (1.76,0.5649) (1.80,0.5568) (1.84,0.5491) (1.88,0.5413) (1.92,0.5338) (1.96,0.5264) (2.00,0.5191)};
\addplot[cE, hzalone, forget plot] coordinates {(0.04,1.0000) (0.08,1.0005) (0.12,1.0045) (0.16,1.0101) (0.20,1.0180) (0.24,1.0246) (0.28,1.0319) (0.32,1.0388) (0.36,1.0457) (0.40,1.0559) (0.44,1.0675) (0.48,1.0762) (0.52,1.0812) (0.56,1.0895) (0.60,1.0981) (0.64,1.1081) (0.68,1.1183) (0.72,1.1287) (0.76,1.1405) (0.80,1.1524) (0.84,1.1646) (0.88,1.1784) (0.92,1.1908) (0.96,1.2049) (1.00,1.2268) (1.04,1.2427) (1.08,1.2572) (1.12,1.2715) (1.16,1.2865) (1.20,1.3013) (1.24,1.3148) (1.28,1.3288) (1.32,1.3424) (1.36,1.3561) (1.40,1.3700) (1.44,1.3825) (1.48,1.3965) (1.52,1.4079) (1.56,1.4194) (1.60,1.4307) (1.64,1.4408) (1.68,1.4508) (1.72,1.4602) (1.76,1.4688) (1.80,1.4773) (1.84,1.4848) (1.88,1.4922) (1.92,1.4991) (1.96,1.5057) (2.00,1.5135)};
\addplot[cE, hzdelta, forget plot] coordinates {(0.04,1.0000) (0.08,0.9995) (0.12,0.9983) (0.16,0.9964) (0.20,0.9931) (0.24,0.9898) (0.28,0.9849) (0.32,0.9782) (0.36,0.9703) (0.40,0.9603) (0.44,0.9487) (0.48,0.9364) (0.52,0.9268) (0.56,0.9160) (0.60,0.9049) (0.64,0.8938) (0.68,0.8814) (0.72,0.8695) (0.76,0.8568) (0.80,0.8440) (0.84,0.8316) (0.88,0.8181) (0.92,0.8059) (0.96,0.7920) (1.00,0.7726) (1.04,0.7639) (1.08,0.7558) (1.12,0.7482) (1.16,0.7398) (1.20,0.7315) (1.24,0.7246) (1.28,0.7163) (1.32,0.7088) (1.36,0.7012) (1.40,0.6931) (1.44,0.6861) (1.48,0.6778) (1.52,0.6717) (1.56,0.6658) (1.60,0.6597) (1.64,0.6544) (1.68,0.6483) (1.72,0.6426) (1.76,0.6370) (1.80,0.6310) (1.84,0.6255) (1.88,0.6197) (1.92,0.6142) (1.96,0.6085) (2.00,0.6016)};
\addplot[cDyn, hzalone, forget plot] coordinates {(0.04,1.0000) (0.08,1.0000) (0.12,1.0000) (0.16,0.9986) (0.20,0.9974) (0.24,0.9978) (0.28,0.9969) (0.32,0.9972) (0.36,0.9991) (0.40,1.0041) (0.44,1.0153) (0.48,1.0245) (0.52,1.0309) (0.56,1.0441) (0.60,1.0572) (0.64,1.0742) (0.68,1.0910) (0.72,1.1087) (0.76,1.1284) (0.80,1.1479) (0.84,1.1682) (0.88,1.1904) (0.92,1.2108) (0.96,1.2340) (1.00,1.2667) (1.04,1.2914) (1.08,1.3131) (1.12,1.3342) (1.16,1.3563) (1.20,1.3784) (1.24,1.3992) (1.28,1.4208) (1.32,1.4424) (1.36,1.4646) (1.40,1.4871) (1.44,1.5080) (1.48,1.5315) (1.52,1.5522) (1.56,1.5733) (1.60,1.5950) (1.64,1.6152) (1.68,1.6362) (1.72,1.6568) (1.76,1.6769) (1.80,1.6974) (1.84,1.7169) (1.88,1.7368) (1.92,1.7563) (1.96,1.7758) (2.00,1.7972)};
\addplot[cDyn, hzdelta, forget plot] coordinates {(0.04,1.0000) (0.08,0.9978) (0.12,0.9983) (0.16,0.9943) (0.20,0.9909) (0.24,0.9866) (0.28,0.9825) (0.32,0.9754) (0.36,0.9661) (0.40,0.9557) (0.44,0.9435) (0.48,0.9316) (0.52,0.9219) (0.56,0.9112) (0.60,0.9007) (0.64,0.8899) (0.68,0.8779) (0.72,0.8663) (0.76,0.8537) (0.80,0.8417) (0.84,0.8295) (0.88,0.8164) (0.92,0.8044) (0.96,0.7902) (1.00,0.7699) (1.04,0.7614) (1.08,0.7532) (1.12,0.7456) (1.16,0.7374) (1.20,0.7287) (1.24,0.7219) (1.28,0.7134) (1.32,0.7057) (1.36,0.6983) (1.40,0.6901) (1.44,0.6829) (1.48,0.6746) (1.52,0.6689) (1.56,0.6638) (1.60,0.6586) (1.64,0.6544) (1.68,0.6497) (1.72,0.6455) (1.76,0.6416) (1.80,0.6376) (1.84,0.6343) (1.88,0.6309) (1.92,0.6277) (1.96,0.6244) (2.00,0.6202)};
\nextgroupplot[title={(c) fast section of A, relative}, ylabel={FDE / FDE$_\mathrm{extrap.}$}, ymode=log, ymin=0.4, ymax=4.5, ytick={0.5,1,2,4}, yticklabels={0.5,1,2,4}]
\addplot[black, thin, densely dashdotted, forget plot] coordinates {(0.04,1.0000) (2.00,1.0000)};
\addplot[gray, thick, densely dotted, forget plot] coordinates {(0.04,1.0000) (0.08,0.9983) (0.12,0.9958) (0.16,0.9923) (0.20,0.9874) (0.24,0.9817) (0.28,0.9735) (0.32,0.9643) (0.36,0.9534) (0.40,0.9404) (0.44,0.9255) (0.48,0.9088) (0.52,0.8955) (0.56,0.8800) (0.60,0.8645) (0.64,0.8492) (0.68,0.8323) (0.72,0.8166) (0.76,0.8000) (0.80,0.7835) (0.84,0.7682) (0.88,0.7516) (0.92,0.7365) (0.96,0.7209) (1.00,0.7021) (1.04,0.6879) (1.08,0.6742) (1.12,0.6613) (1.16,0.6483) (1.20,0.6357) (1.24,0.6239) (1.28,0.6119) (1.32,0.6006) (1.36,0.5895) (1.40,0.5785) (1.44,0.5685) (1.48,0.5581) (1.52,0.5488) (1.56,0.5398) (1.60,0.5310) (1.64,0.5228) (1.68,0.5147) (1.72,0.5069) (1.76,0.4993) (1.80,0.4919) (1.84,0.4848) (1.88,0.4777) (1.92,0.4709) (1.96,0.4642) (2.00,0.4573)};
\addplot[cE, hzalone, forget plot] coordinates {(0.04,1.0000) (0.08,1.0165) (0.12,1.0416) (0.16,1.0666) (0.20,1.0992) (0.24,1.1261) (0.28,1.1585) (0.32,1.1903) (0.36,1.2210) (0.40,1.2577) (0.44,1.2914) (0.48,1.3252) (0.52,1.3460) (0.56,1.3710) (0.60,1.3969) (0.64,1.4205) (0.68,1.4476) (0.72,1.4738) (0.76,1.5012) (0.80,1.5310) (0.84,1.5581) (0.88,1.5876) (0.92,1.6165) (0.96,1.6471) (1.00,1.6862) (1.04,1.7113) (1.08,1.7361) (1.12,1.7618) (1.16,1.7880) (1.20,1.8146) (1.24,1.8386) (1.28,1.8631) (1.32,1.8870) (1.36,1.9099) (1.40,1.9325) (1.44,1.9522) (1.48,1.9720) (1.52,1.9884) (1.56,2.0032) (1.60,2.0171) (1.64,2.0284) (1.68,2.0390) (1.72,2.0483) (1.76,2.0558) (1.80,2.0627) (1.84,2.0678) (1.88,2.0722) (1.92,2.0757) (1.96,2.0783) (2.00,2.0818)};
\addplot[cE, hzdelta, forget plot] coordinates {(0.04,1.0000) (0.08,1.0013) (0.12,1.0030) (0.16,1.0039) (0.20,1.0043) (0.24,1.0027) (0.28,0.9999) (0.32,0.9949) (0.36,0.9879) (0.40,0.9787) (0.44,0.9673) (0.48,0.9540) (0.52,0.9413) (0.56,0.9272) (0.60,0.9127) (0.64,0.8983) (0.68,0.8829) (0.72,0.8679) (0.76,0.8528) (0.80,0.8377) (0.84,0.8235) (0.88,0.8088) (0.92,0.7952) (0.96,0.7812) (1.00,0.7642) (1.04,0.7545) (1.08,0.7455) (1.12,0.7369) (1.16,0.7284) (1.20,0.7202) (1.24,0.7129) (1.28,0.7055) (1.32,0.6984) (1.36,0.6915) (1.40,0.6845) (1.44,0.6781) (1.48,0.6712) (1.52,0.6652) (1.56,0.6591) (1.60,0.6530) (1.64,0.6472) (1.68,0.6409) (1.72,0.6347) (1.76,0.6283) (1.80,0.6218) (1.84,0.6153) (1.88,0.6086) (1.92,0.6019) (1.96,0.5949) (2.00,0.5874)};
\addplot[cDyn, hzalone, forget plot] coordinates {(0.04,1.0000) (0.08,1.0083) (0.12,1.0120) (0.16,1.0152) (0.20,1.0160) (0.24,1.0151) (0.28,1.0109) (0.32,1.0056) (0.36,1.0005) (0.40,0.9951) (0.44,0.9935) (0.48,0.9919) (0.52,0.9899) (0.56,0.9977) (0.60,1.0084) (0.64,1.0248) (0.68,1.0451) (0.72,1.0691) (0.76,1.0977) (0.80,1.1303) (0.84,1.1663) (0.88,1.2065) (0.92,1.2496) (0.96,1.2969) (1.00,1.3536) (1.04,1.4043) (1.08,1.4557) (1.12,1.5097) (1.16,1.5657) (1.20,1.6235) (1.24,1.6806) (1.28,1.7394) (1.32,1.7992) (1.36,1.8587) (1.40,1.9188) (1.44,1.9767) (1.48,2.0353) (1.52,2.0908) (1.56,2.1447) (1.60,2.1976) (1.64,2.2475) (1.68,2.2963) (1.72,2.3431) (1.76,2.3873) (1.80,2.4301) (1.84,2.4700) (1.88,2.5081) (1.92,2.5442) (1.96,2.5785) (2.00,2.6129)};
\addplot[cDyn, hzdelta, forget plot] coordinates {(0.04,1.0000) (0.08,1.0004) (0.12,0.9997) (0.16,0.9984) (0.20,0.9958) (0.24,0.9921) (0.28,0.9871) (0.32,0.9800) (0.36,0.9718) (0.40,0.9616) (0.44,0.9501) (0.48,0.9380) (0.52,0.9277) (0.56,0.9161) (0.60,0.9047) (0.64,0.8940) (0.68,0.8827) (0.72,0.8725) (0.76,0.8626) (0.80,0.8533) (0.84,0.8452) (0.88,0.8373) (0.92,0.8306) (0.96,0.8235) (1.00,0.8150) (1.04,0.8135) (1.08,0.8128) (1.12,0.8129) (1.16,0.8132) (1.20,0.8140) (1.24,0.8164) (1.28,0.8190) (1.32,0.8227) (1.36,0.8268) (1.40,0.8312) (1.44,0.8361) (1.48,0.8414) (1.52,0.8479) (1.56,0.8552) (1.60,0.8627) (1.64,0.8708) (1.68,0.8790) (1.72,0.8878) (1.76,0.8967) (1.80,0.9057) (1.84,0.9149) (1.88,0.9241) (1.92,0.9333) (1.96,0.9424) (2.00,0.9516)};
\end{groupplot}
\node[anchor=south] at ($(group c2r1.north)+(0,0.42cm)$) {\pgfplotslegendfromname{hzlegend}};
\end{tikzpicture}
\caption{Position error along the horizon (dashed: host alone, solid: with \dlt). The vertical lines mark the horizons of Table~\ref{tab:main}.}
\label{fig:horizon}
\end{figure*}

\subsection{Error over the Horizon}
\label{sec:horizon}

Fig.~\ref{fig:horizon} shows the error along the horizon, for the DDM protocol and for the fast section of recording A. All predictors start at the same state. After one step, they all have the same error of 3\,cm, because the first position depends only on the measured start velocities. These 3\,cm therefore come from a mismatch between the position and the velocity measurements. Beginning with the second step, the physics model falls behind the extrapolation. As expected for an acceleration bias integrated twice, the distance grows faster than the horizon, from 2\,cm at 0.6\,s to 13\,cm at 1.2\,s and 50\,cm at 2\,s. This effect is even stronger in the fast section (Fig.~\ref{fig:horizon}c), where the FDE of the physics model is about twice as high as the extrapolation FDE. With \dlt, it is 72\pc lower and 41\pc below extrapolation, out largest gain on real data.

Most of this bias gets erased by \dlt. Up to 0.6\,s, the physics model with \dlt stays within 3\pc of the oracle and slowly drifts apart afterwards. The error stays below the extrapolation (60\pc of it at 2\,s, Fig.~\ref{fig:horizon}b).

Which host is ahead also changes with the horizon. Without \dlt, \dyn beats the physics model at 0.6\,s and is only 6\pc worse than extrapolation. After 2\,s, however, it is worse by 80\pc and behind the physics model. With \dlt, the physics model and \dyn are within 1\pc of each other at 0.6\,s and within 3\pc at 2\,s. At 2\,s in the fast section (Fig.~\ref{fig:horizon}c), the physics model and \dyn with \dlt are at 59 and 95\pc of extrapolation, respectively. \dlt starts empty at the start of this section for the physics model and of the drive for \dyn (at most 0.0025 relative error). On this dataset, a horizon of 0.6\,s is therefore too short to tell the models apart. \dlt removes the error part that grows with the horizon, which matters for an MPC over 2\,s.

\subsection{Closed-Loop Simulation}
\label{sec:sim}

\begin{table}[t]
\caption{Closed-loop simulation with $\eta = 0.85$ and three runs per entry. Each change refers to the same host without \dlt.}
\label{tab:sim}
\centering\scriptsize
\setlength{\tabcolsep}{2.6pt}
\begin{tabular}{@{}lccccc@{}}
\toprule
 & \multicolumn{2}{c}{Ovals} & \multicolumn{3}{c}{Road courses} \\
\cmidrule(lr){2-3}\cmidrule(lr){4-6}
 & LVMS & IMS & Putnam & Suzuka & Yas M. \\
\midrule
\multicolumn{6}{@{}l}{\emph{Physics model: lap time [s]}}\\
\quad with \dlt & $-$0.26 & $-$0.98 & $-$0.40 & $-$0.30 & $-$1.59 \\
\quad with offline table & +0.17 & +0.84 & $-$0.46 & $-$0.21 & $-$0.91 \\
\multicolumn{6}{@{}l}{\emph{Time off track (share of driving time), alone $\to$ with \dlt{} [\%]}}\\
\quad Physics & 3.5 $\to$ 0 & 11.0 $\to$ 0 & 0 $\to$ 0 & 0.7 $\to$ 0.3 & 0.6 $\to$ 0.2 \\
\quad \ddmstep & 9.5 $\to$ 0 & 16.4 $\to$ 0 & 0 $\to$ 0 & 0 $\to$ 0 & 0.4 $\to$ 0.3 \\
\quad \dyn & 0 $\to$ 0 & 5.0 $\to$ 0 & 0 $\to$ 0 & 0 $\to$ 0 & 0.2 $\to$ 0.2 \\
\quad Physics, offline table & 25.7 & 28.4 & 0 & 0.5 & 0.2 \\
\multicolumn{6}{@{}l}{\emph{Change of the prediction error at 2\,s with \dlt [\%]}}\\
\quad Physics & $-$19 & $-$57 & $-$12 & $-$17 & $-$16 \\
\quad \ddmstep & $-$82 & $-$78 & $-$1 & $-$5 & +1 \\
\quad \dyn & $-$57 & $-$59 & +7 & $-$3 & +10 \\
\bottomrule
\end{tabular}
\end{table}

Lastly, we tested \dlt in a closed-loop simulation based on Open Car Dynamics~\cite{ocd} with parameters fitted to BETTY data. For this test, the three hosts were trained again and the offline table was filled on one hour of simulated driving on Spielberg, Monza and Putnam, while \dlt was left as is. As DDM's evaluation script runs only on recordings, this test uses our own implementation \ddmstep (history 15), built before we switched to that script; like \dyn, it queries the network at every step instead of holding the coefficients. A sampling-based MPC~\cite{williams2017} with a contouring cost~\cite{liniger2015} evaluates 1\,024 candidates over 2\,s in each cycle and may use a share $\eta$ of the friction envelope. The layouts of Putnam and LVMS come from the recorded drives, all others from~\cite{tumdb}. The ovals are banked, LVMS with the measured profile (5.9 to 21.1$^\circ$) and IMS with 9.2$^\circ$ in the turns. All hosts hold the bank angle at the current position over the horizon, and the networks again use it only in the kinematics. The plant, however, integrates its heading on a flat plane and receives the bank only through lateral and normal forces, while all hosts use $\dot\psi = r/\cos\theta$. This mismatch of up to 7\pc in the LVMS turns and 1.3\pc on IMS affects all hosts equally and is invisible to \dlt.

Table~\ref{tab:sim} shows the results; each prediction error is taken on the path driven with the respective variant. The physics model predicts better with \dlt on each track by 12 to 57\pc, even with exact measurements. The networks already do well on road courses, therefore \dlt changes little there and even makes \dyn slightly worse (+7 and +10\pc). Meanwhile, on unseen ovals, their errors are larger, of which \dlt removes 57 to 82\pc. Without \dlt, the car leaves the track in five out of six host--oval combinations. With \dlt, it never leaves the ovals. On the road courses, \dlt reduced the lap time of the physics model by 0.2 to 1.0\pc, on LVMS by 0.7\pc. On IMS, the gain of 2.0\pc mainly reflects that the host alone is off track for 11\pc of the time. On the road courses, the offline table predicts as well as the online table, but it fails on the ovals. With a sudden grip drop of 30\pc at $\eta = 0.7$, the online table predicted 6 to 7\pc better than the offline table after the drop and 2 to 7\pc worse before it. At $\eta = 0.85$, every run left the track.

The gain in lap time depends on how aggressively the controller drives and is almost gone at $\eta = 0.7$. On Monza at the full friction limit, \dlt was slightly slower than the host alone, and the car stopped after a chicane in three of five starts, with \ddmstep and \dlt in four of five. With the physics host, these stops vanished with the correction faded out below 15\,m/s, a threshold chosen on these runs.

\section{Conclusion}

We showed that the error which a car observes one sample later is enough to correct a large part of the prediction error on an unknown track. It works with a physics model, DDM and \dyn without changing the hosts themselves and brings a single-track model close to the best learned host (about 8\pc behind with neutral initial values). In the closed-loop simulation, the car stayed on track on ovals unknown to the models. A fixed table predicted as well on road courses like the training tracks, but failed on the ovals and was 6 to 7\pc worse after a sudden grip drop. There is little to correct when the host already fits the track, and for short horizons the accuracy can even get worse. All cross-track tests on real data target one track, LVMS, and the closed loop was only simulated. The hyperparameters of \dlt were chosen with another LVMS recording, the fade-out threshold on the test runs of the simulation. \ddms stays above extrapolation at 2\,s even with \dlt. Filling the table takes 20\,s; until then, only the fast term corrects. Cells that distinguish the turn direction did not help during development. Next, we want to adapt controller limits together with the model.

\section*{Acknowledgment}
The author thanks the teams behind the RACECAR and BETTY datasets and the authors of Deep Dynamics for sharing their data and code. Claude (Anthropic) assisted with editing and code. The author checked all content and is fully responsible for it.

\end{document}